\documentclass[10pt,twocolumn,letterpaper]{article}
\usepackage{iccv}
\usepackage{times}
\usepackage{epsfig}
\usepackage{graphicx}
\usepackage{amsmath}
\usepackage{amssymb}

\usepackage[breaklinks=true,bookmarks=false]{hyperref}

\iccvfinalcopy 

\begin{document}

\title{Learning RGB-D Salient Object Detection using background enclosure, depth contrast, and top-down features}

\author{
Riku Shigematsu~~~~~~~~~~~~David Feng\\
Australian National University\\
{\tt\small riku.research@gmail.com}\\
{\tt\small david.feng@data61.csiro.au}
\and
Shaodi You~~~~~~~~Nick Barnes\\
Australian National University\\
Data61-CSIRO\\
{\tt\small shaodi.you@data61.csiro.au}\\
{\tt\small nick.barnes@data61.csiro.au}
}

\maketitle

\begin{abstract}
   Recently, deep Convolutional Neural Networks (CNN) have demonstrated strong performance on RGB salient object detection. Although, depth information can help improve detection results, the exploration of CNNs for RGB-D salient object detection remains limited. Here we propose a novel deep CNN architecture for RGB-D salient object detection that exploits high-level, mid-level, and low level features. Further, we present novel depth features that capture the ideas of background enclosure and depth contrast that are suitable for a learned approach. We show improved results compared to state-of-the-art RGB-D salient object detection methods. We also show that the low-level and mid-level depth features both contribute to improvements in the results. Especially, F-Score of our method is 0.848 on RGBD1000 dataset, which is 10.7\% better than the second place.
\end{abstract}

\section{Introduction}

In computer vision, visual saliency attempts to predict which parts of an image attract human attention.
Saliency can be used in the context of many computer vision problems such as compression~\cite{compression_saliency}, object detection \cite{LuoCVPR2014}, visual tracking~\cite{Mahadevan_visualtarget_2009}, and retargeting images and videos~\cite{retargeting}. 
Early work attempted to predict human gaze direction on images \cite{1998saliency}.
However, in recent years, the field has focused on salient object detection, finding salient objects or regions in an image (e.g., \cite{2009saliency,2015saliency}). 
Rather than attention, the main task for salient object detection is to produce a fine grained binary mask for salient objects, based on human annotation.

\begin{figure}
\begin{center}
\includegraphics[width=\linewidth]{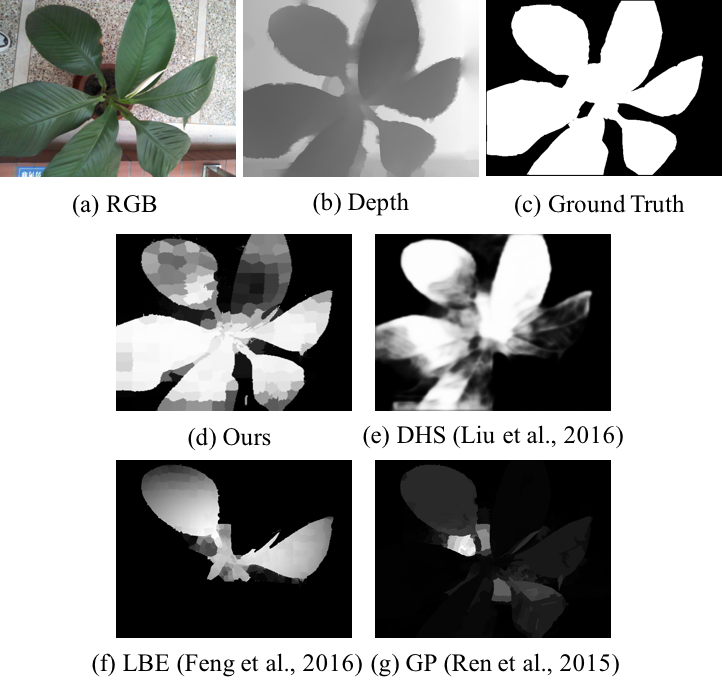}
\end{center}
   \caption{Comparing our RGB-D salient object detector output with other salient object detection methods.}
\label{fig:long}
\label{fig:onecol}
\end{figure}

Most salient object detection methods are based on RGB images. However, depth plays a strong role in human perception, and it has been shown that human perception of salient objects is also influenced by depth~\cite{Lang_depth_matters}. Thus, RGB-D salient object detection methods have been proposed~\cite{LBE,NJUDS,peng2014rgbd,RGBDCNN,Ren_2015_CVPR_Workshops} and have demonstrated superior performance in comparison to RGB-only methods.

While many salient object detection methods adopt a bottom-up strategy \cite{LBE,Guorgbd2015,NJUDS,peng2014rgbd,Ren_2015_CVPR_Workshops}, recently, top-down methods through machine learning have demonstrated superior performance \cite{Lee_2016_CVPR,Liu_2016_CVPR,RGBDCNN,Zhao_2015_CVPR}.
Recent papers have tackled top-down learning for RGB salient object detection using deep CNN for their methods~\cite{Lee_2016_CVPR,Liu_2016_CVPR,Zhao_2015_CVPR}. However, it is not yet clear whether deep CNNs are
effective for RGB-D saliency detection. 

The approach of this paper is premised on observations of the performance of state-of-the-art approaches in salient object detection. Top-down information has been shown to be effective in RGB salient object detection (top-down information also plays a role in human visual focus of attention \cite{1998saliency}). Further, in RGB-D salient object detection, the effectiveness of background enclosure and of depth contrast have been demonstrated. Finally, deep CNNs have been shown to be effective for RGB salient object detection.

This paper makes three major contributions. (1) We propose a novel learning architecture that provides the first complete RGB-D salient object detection systems using a deep CNN, incorporating high level features, depth contrast and low-level features, and a novel mid-level feature. (2) We introduce the background enclosure distribution, BED, a novel mid-level depth feature that is suitable for learning based on the idea of background enclosure. (3) We introduce a set of low level features that are suitable for learning that incorporate the idea of depth contrast. 

We show that our new approach produces state-of-the-art results for RGB-D salient object detection. 
Further, we evaluate the effectiveness of adding depth features, and of adding the mid-level feature in particular. In ablation studies, we show that incorporating low-level features that incorporate depth contrast performs better than RGB saliency alone, and that adding our new mid-level feature, BED, improves results further.

The rest of the paper is organized as follows:
Section 2 presents the related work on salient object detection and deep CNN methods for the other computer vision task.
In Section 3, we introduce the idea on RGB-D salient object detection with a deep CNN architecture and our algorithms for salient object detection. This is followed by the detailed neural network structure, technical details and the training approach in Section 4. We introduce systematic and extensive experimental results in Section 5, and conclude in Section 6.

\section{Related Work}

Saliency detection to model eye movements began with low-level hand-crafted features, with classic work by Itti \etal~\cite{1998saliency} being influential. A variety of salient object detection methods have been proposed in recent years, we focus on these as more relevant to our work.

\paragraph{RGB Salient object detection}
In RGB salient object detection, methods often measure constrast between features of a region versus its surrounds, either locally and/or globally \cite{Cheng_2011_CVPR,1998saliency}. Contrast is mostly computed with respect to appearance-based features such as colour, texture, and intensity edges \cite{cheng_2013_ICCV,Jiang_2013_CVPR}.

\paragraph{RGB salient object detection using deep CNNs}
Recently, methods using deep CNNs have obtained strong results for RGB salient object detection. Wang \etal \cite{Wang_2015_CVPR} combine local information and a global search. 
Often the networks make use of deep CNN networks for object classification for a large number of classes, specifically VGG16~\cite{VGG16} or GoogleNet~\cite{GoogleNet}. 
Some utilize these networks for extracting the low features~\cite{Lee_2016_CVPR,Li_2016_CVPR,Liu_2016_CVPR}. 
Lee \etal incorporate high-level features based on these networks, along with low level features~\cite{Lee_2016_CVPR}. 
This approach to incorporating top-down semantic information about objects into salient object detection has been effective.

\paragraph{RGB-D Salient Object Detection}
Compared to RGB salient object detection, fewer methods use RGB-D values for computing saliency. Peng \etal calculate a saliency map by combining low, middle, and high level saliency information~\cite{peng2014rgbd}. 
Ren \etal calculate region contrast and use background, depth, and orientation priors. They then produce a saliency map by applying PageRank and a MRF to the outputs~\cite{Ren_2015_CVPR_Workshops}. Ju \etal calculate the saliency score using anisotropic center-surround difference and produce a saliency map by refining the score applying Grabcut segmentation and a 2D Gaussian filter~\cite{NJUDS}. Feng \etal 
improve RGB-D salient object detection results based on the idea that salient objects are more likely to be in front of their surroundings for a large number of directions~\cite{LBE}.
All the existing RGB-D methods use hand-crafted parameters, such as for scale and weights between metrics.
However, real world scenes contain unpredictable object arrangements for which fixed hand coded parameters may limit generalization.
No published papers have yet presented a CNN architecture for RGB-D salient object detection. A preliminary paper (Arxiv only) uses only low-level color and depth features~\cite{RGBDCNN}. 

\paragraph{Datasets}
Two datasets are widely used for RGB-D salient object detection, RGBD1000~\cite{peng2014rgbd} and NJUDS2000~\cite{NJUDS}. 
The RGBD1000 datasets contain 1000 RGB-D images captured by a standard Microsoft Kinect. The NJUDS2000 datasets contain around 2000 RGB-D images captured by a Fuji W3 stereo camera.

\section{A novel deep CNN architecture for detecting salient objects in RGB-D images}

\begin{figure*}[t]
\begin{center}
\includegraphics[width=\linewidth]{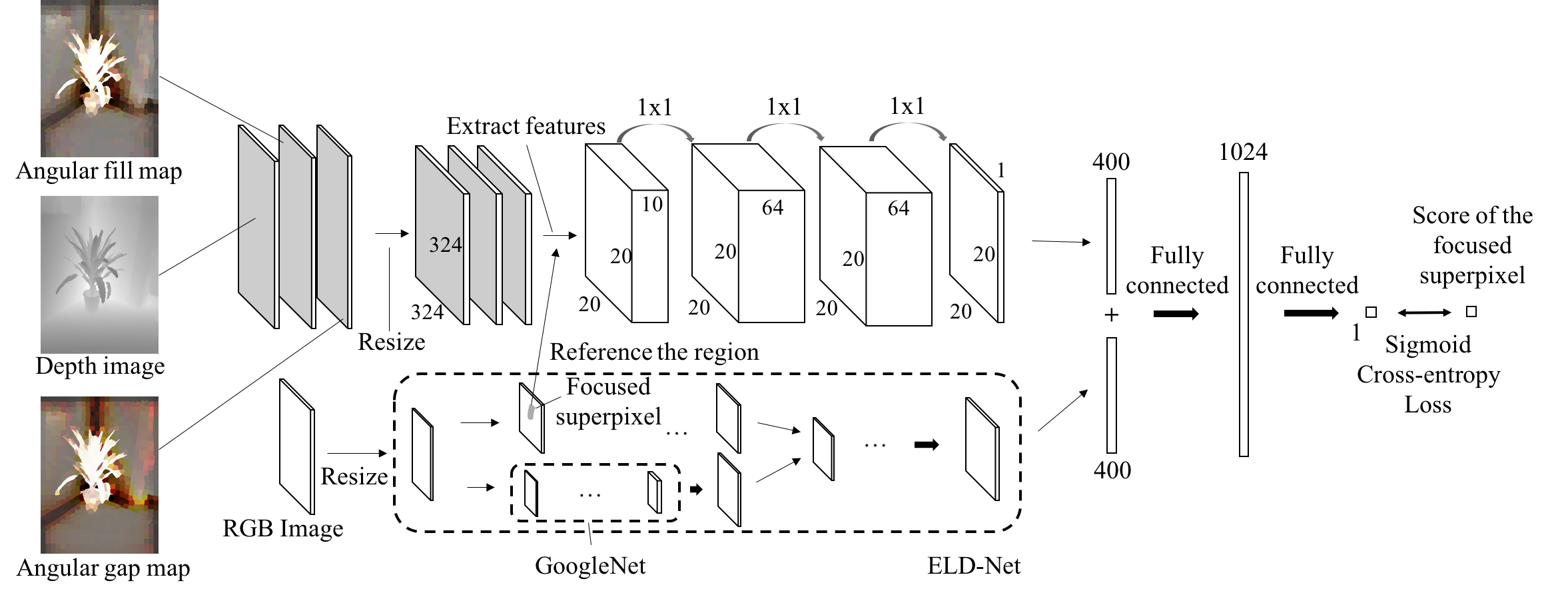}
\end{center}
   \caption{The whole architecture of our method. We extract ten superpixel-based handcrafted depth features for inputs (Section 3.1 and 3.2). Then we combine the depth features by concatenating the output with RGB low-level and high-level saliency features output (Section 3.3 and 3.4). Finally, we compute the saliency score with two fully connected layers. }
\label{whole_architecture}
\end{figure*}

In this section, we introduce our approach to RGB-D salient object detection. Our novel deep CNN learning architecture is depicted in Figure \ref{whole_architecture}. We combine the strengths of previous approaches to high-level and low-level feature-based deep CNN RGB salient object detection \cite{Lee_2016_CVPR}, with a depth channel, incorporating raw depth, low level cues to capture depth contrast, and a novel BED feature to capture background enclosure. 

\subsection{BED Feature}

\begin{figure}[t]
\begin{center}
\includegraphics[width=0.9\linewidth]{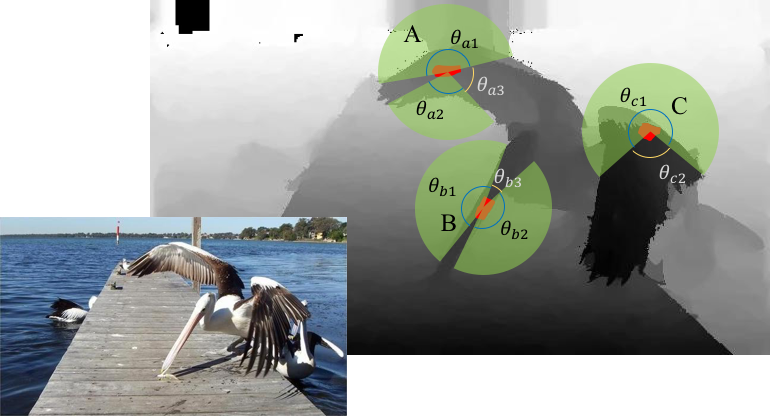}
\end{center}
   \caption{The concepts of the foreground function $f(P,t)$ and the opposing background function $g(P,t)$. For example, $f(P,t)=\frac{\theta_{a1} + \theta_{a2}}{2\pi}$ and $g(P,t)=\frac{\theta_{a3}}{2\pi}$ at point A.}
\label{develop_input_image}
\end{figure}

High-level and low-level features have been shown to lead to high performance for detecting salient objects in RGB images in a deep CNN framework \cite{Lee_2016_CVPR}. We also know that  the effective encoding of depth input can improve convergence and final accuracy where training data is limited~\cite{Gupta2014}. Here we add a novel mid-level feature that aims to represent the depth enclosure of salient regions for a learning approach, called the Background Enclosure Distribution (BED) . BED relies on learning rather than hand-coded parameters that limit generalization. 
 
Our proposed BED feature captures the enclosure distribution properties of a patch, that is, the spread of depth change in the surrounds, based on the idea that salient objects are more likely to be in front of their surroundings in a large number of directions. 
BED is inspired by LBE for salient object detection, which has been shown to be an effective hand-crafted feature for non-learned salient object detection~\cite{LBE}.

For each superpixel $P$, we define a foreground function $f(P,t)$ that measures the spread of directions (the integral over angle) in which P is in front of its background set, consisting of all patches with greater depth than $depth(P)+t$. That is, there is no superpixel in front of it in a particular direction, and at least one has greater depth. We also define an opposing background function $g(P,t)$ that measures the size of the largest angular region in which the superpixel is not in front of its background set.

We aim to measure the distribution of $f$ and $g$ over a range of distance to the background (i.e., $t$) to provide a stable representation of background enclosure. The distribution functions are given by:

\begin{eqnarray}
  F(P, a, b) = \int_a^b f(P,B(P,t))dt  \\
  G(P, c, d) = \int_c^d 1-g(P,B(P,t))dt,
\end{eqnarray}
where $(a,b)$ and $(c,d)$ are some range of depth.
We define a quantization factor $q$ over the total range of depth of interest.
Our BED feature consists of two distribution sets $FF$ and $GG$:
\begin{eqnarray}
  FF(P,\sigma,q) = \left\{F(P,r,r - \sigma/q) | r\in\left\{\sigma/q, 2\sigma/q, ..., \sigma\right\}  \right\}\\
  GG(P,\sigma,q) = \left\{G(P,r,r - \sigma/q) | r\in\left\{\sigma/q, 2\sigma/q, ..., \sigma\right\}  \right\}.
\end{eqnarray}
This provides a rich representation of image structure that is descriptive enough to provide strong discrimination between salient and non salient structure.

We construct a $20 \times 20$ feature layer for each of these distribution slices. This results in $2q$ feature layers for our BED feature. 

\subsection{Low-level Depth Features}
\begin{figure}[t]
\begin{center}
\includegraphics[width=0.9\linewidth]{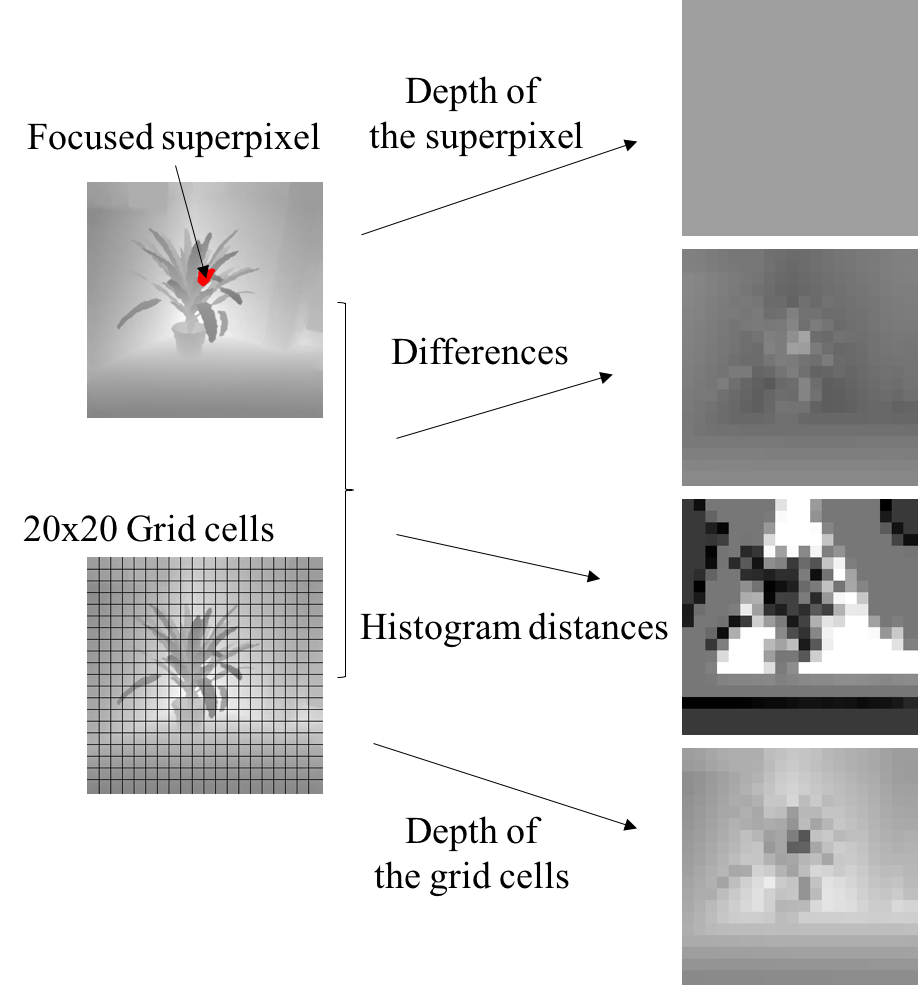}
\end{center}
   \caption{Our four $20 \times 20$ depth feature layers.}
\label{fig:extract_depth_low}
\end{figure}

In addition to background enclosure, we also capture the idea of depth contrast, that has been shown to be effective in previous work~\cite{NJUDS,RGBDCNN,Ren_2015_CVPR_Workshops}. 
In addition to six BED features, we extract four low-level depth features from every superpixel.
The extracted features are illustrated in Table \ref{table:low_features} and Figure \ref{fig:extract_depth_low}.

\begin{table}
\begin{center}
\begin{tabular}{|c|c|}
\hline
Depth feature name & The number of the features \\
\hline\hline
Depth of focused superpixel & 1 \\
\hline
Depth of the grid pixel & 1 \\
\hline
Depth contrast & 1 \\
\hline
Histogram distance & 1\\
\hline
Angular density components & 3\\
\hline
Angular gap components & 3\\
\hline
\end{tabular}
\end{center}
\caption{The depth features extracted from the focused superpixel and a grid cell.}
\label{table:low_features}
\end{table}

We use the SLIC algorithm~\cite{SLIC} on the RGB image to segment it into superpixels (approximately $18 \times 18$ superpixels per image). In every learning step, we focus on one superpixel, calculate how salient the superpixel will be, compare it with ground truth, and perform back propagation.

For every focused superpixel, we calculate the average depth value to form a $20 \times 20$ layer of these values. We also subdivide the image into $20 \times 20$ grid cells and calculate the average value for each to form a $20 \times 20$ layer. To capture depth contrast (local and global) that has been shown to be effective in RGB-D saliency, we create a $20 \times 20$ contrast layer between the depth of the superpixels and grid cells. We compute the contrast layer simply by subtracting the average depth value of each grid cell from the average depth value for each superpixel. Finally, we calculate the difference between the depth histogram of the focused superpixel and grid cells. We divide the entire range of depth values into 8 intervals and make the histogram of the distribution of the depth values of each superpixel and grid cell. To measure histogram contrast, we calculate the $\chi^2$ distance between focused superpixel and the grid pixel features. The equation is illustrated in Equation (\ref{eqn:ChiSquare}):
\begin{equation}
f(x,y) = \frac{1}{2} \sum_{i=1}^8 \frac{(x_i-y_i)^2}{(x_i+y_i)},
\label{eqn:ChiSquare}
\end{equation}
where $x_i$ is the number of depth values in quanta $i$ for the superpixel, and $y_i$ is the number of depth values in the range $i$ for the grid cell.
These features are also inspired by the RGB features that are shown to be effective in the original version of ELD-Net~\cite{Lee_2016_CVPR}.

\subsection{RGB low and high level saliency from ELD-Net}
To represent high-level and low-level features for RGB, we make use of the extended version of ELD-Net~\cite{Lee_2016_CVPR}. We choose ELD-Net because this method is a state-of-the-art RGB saliency method and the network architecture is easy to extend to RGB-D saliency. From personal correspondence, Lee \etal published the source code for a better performing method in https://github.com/gylee1103/ELDNet. Rather than using VGG-Net as per the ELD-Net paper, this version uses GoogleNet~\cite{GoogleNet} to extract high level features, and does not incorporate all low-level features.

\subsection{Non-linear combination of depth features}

The $20 \times 20$ depth images, the other low-level feature maps that capture depth contrast, and the BED feature map, as described in Section 3.1 and 3.2, need to be combined to capture background enclosure, depth contrast, and absolute depth. We also incorporate RGB information from the focused superpixel. In order to capture non-linear combinations of these, we use three convolutional layers followed by a fully convolutional layer to form our depth output.

\subsection{Concatenation of Color and Depth Features}
In order to effectively exploit color features, we make use of the pretrained caffemodel of ELD-Net~\cite{Lee_2016_CVPR} to initialize the weights of color features. The calculated $1 \times 20 \times 20$ color feature layers are concatenated with the depth feature output as shown in the Figure \ref{whole_architecture}. 

We then connect the $1 \times 20 \times 20 + 1 \times 20 \times 20$ concatenated output features with a fully connected layer and calculate the saliency score for the focused superpixel. We calculate the cross entropy loss for a softmax classifier to evaluate the outputs. The cross entropy loss is calculated as follows:
\begin{equation}
E = -\bigl\{ p \log \hat{p} + (1-p) \log (1-\hat{p})\bigl\},
\end{equation}
where $p$ is the calculated saliency score of the focused superpixel and $\hat{p}$ is the average saliency score for the ground truth image.

\section{RGB-D saliency detection system}
We develop our learning architecture for salient object detection based on the Caffe~\cite{Caffe} deep learning framework. For faster learning, our training uses CUDA on a GPU.

\begin{figure}[t]
\begin{center}
\includegraphics[width=0.9\linewidth]{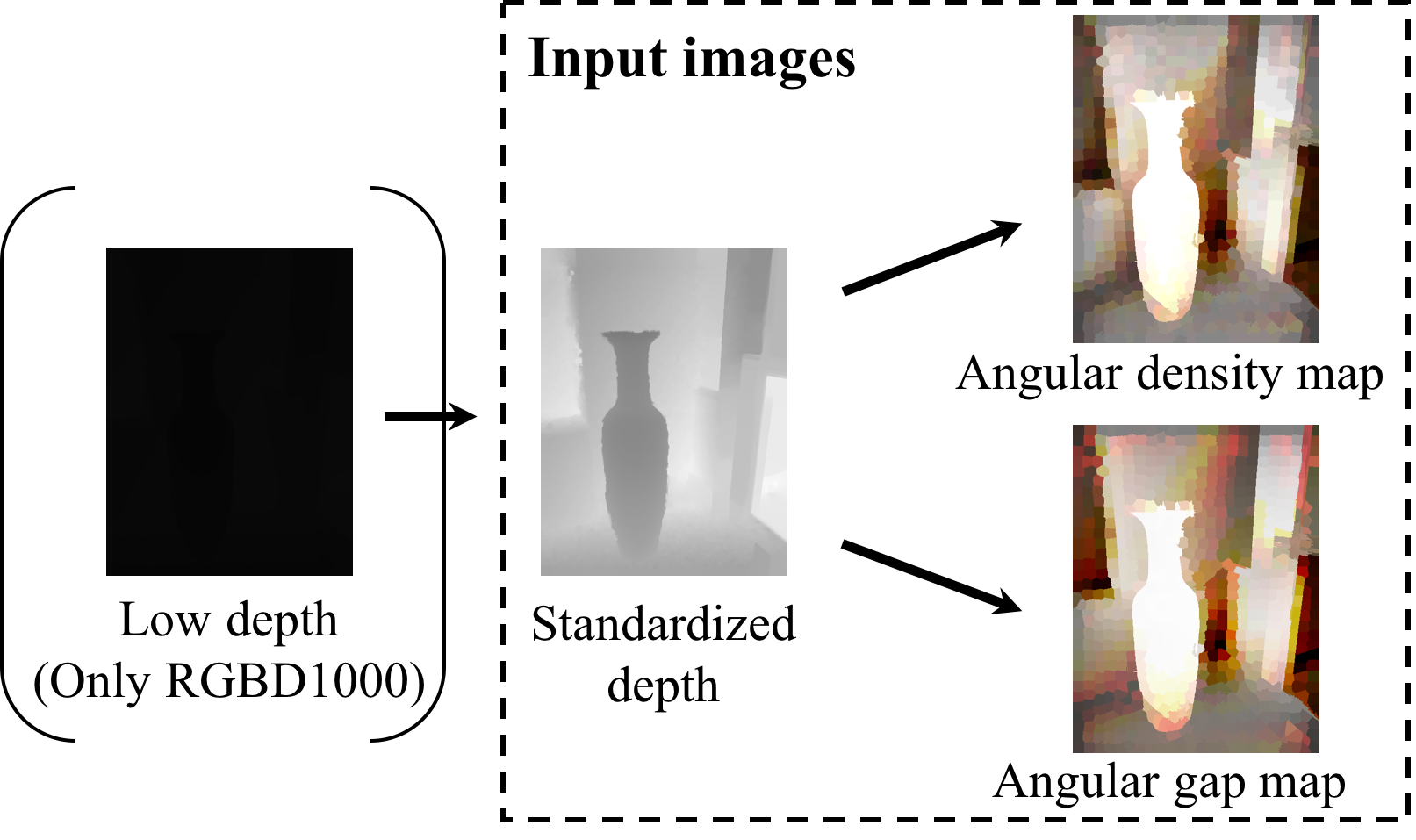}
\end{center}
   \caption{Developing input images from a depth image.}
\label{develop_input_image}
\end{figure}

\begin{figure}[t]
\begin{center}
\includegraphics[width=0.9\linewidth]{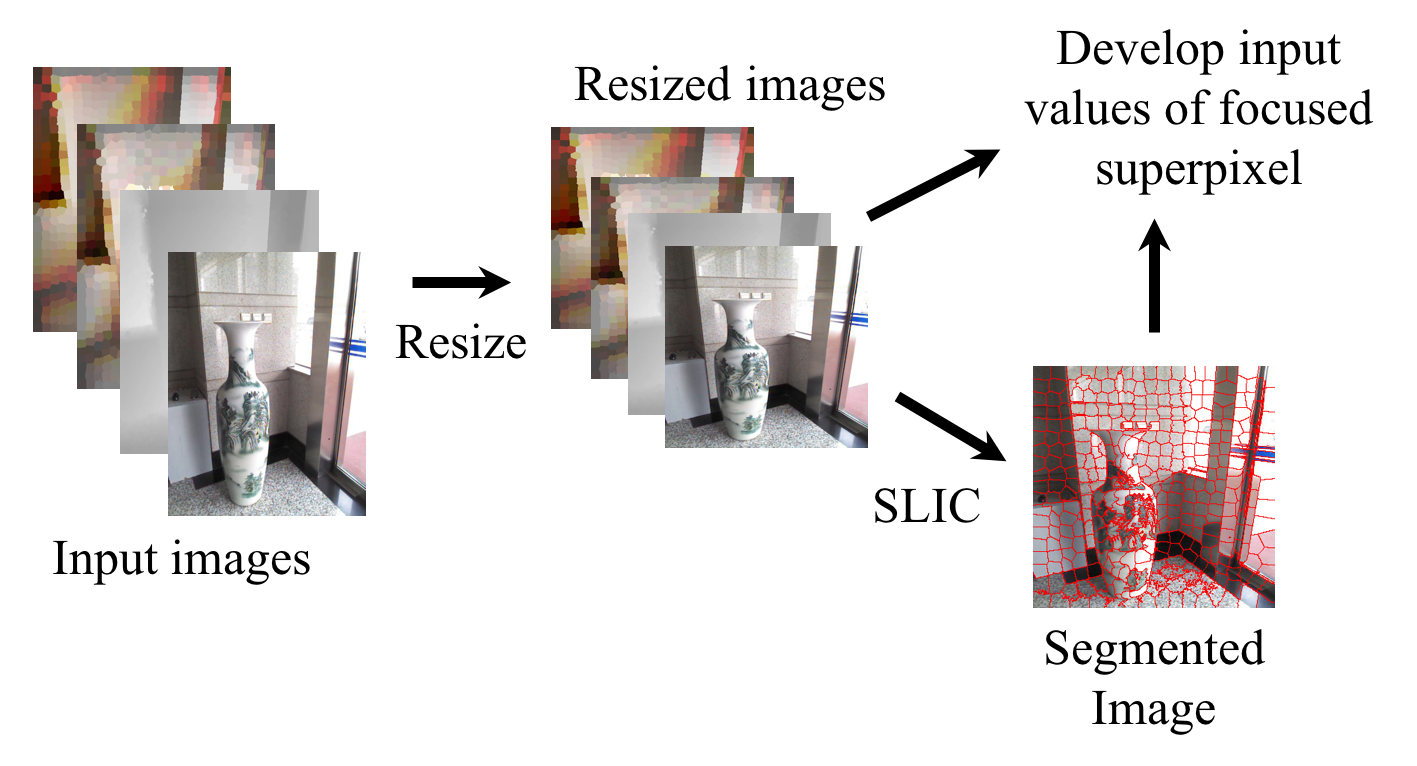}
\end{center}
   \caption{Extracting feature values of the focused superpixel from various input images.}
\label{extract_features}
\end{figure}

\subsection{Preprocessing on depth and color images}
Since we concatenate the color and depth values, we want to synchronize the scale of depth values with color values. Hence, if required, we normalize the depth value to the same scale, i.e., 0 to 255, before extracting depth features. Depth values of RGBD1000~\cite{peng2014rgbd} are represented with greater bit depth and so require normalization. On NJUDS2000~\cite{NJUDS} the scale of depth values are already 0 - 255, and so are not modified.
After normalization, we resize the color and depth images to $324 \times 324$.

\subsection{Superpixel Segmentation}
We use gSLICr~\cite{gSLICr_2015}, the GPU version of SLIC, to segment the images into superpixels. We divide each image into approximately $18 \times 18$ superpixels, following Lee \etal~\cite{Lee_2016_CVPR}. Note that gSLICr may combine small superpixels with nearby superpixels~\cite{gSLICr_2015}.

\subsection{Extracting low-level depth features}

Following this, we create four $20 \times 20$ layers from each of the low-level depth features. The first consists of the average value of the spatially corresponding focused superpixel for each of the $20 \times 20$ inputs; the second is composed from the average depth values of $20 \times 20$ grid cells; the third layer consists of the difference of depth values between the mean depth of the focused superpixel and the mean depth of each of the grid cells; and the last layer consists of the histogram distance between the superpixel and grid cells. Figure \ref{fig:extract_depth_low} illustrates this process.

\subsection{Extracting BED features}
In order to calculate BED efficiently, we pre-compute the angular density components and angular fill components. Three channels are computed for each of equation (2) and (3), where $q=3$ over the intervals between $0,\frac{\sigma}{3}$, $\frac{2\sigma}{3}$, $\sigma$ where $\sigma$ is the standard deviation of the mean patch depths. The calculated values are connected to our architecture in the same way as loading color images. For each focused superpixel, we calculate each BED feature, for a total of six $20 \times 20$ feature maps.
These are concatenated with depth to form a $(4+6)\times 20 \times 20$ feature input for each focused super pixel.

\subsection{Improving the learning rate}

\begin{figure}[t]
\begin{center}
\includegraphics[width=0.9\linewidth]{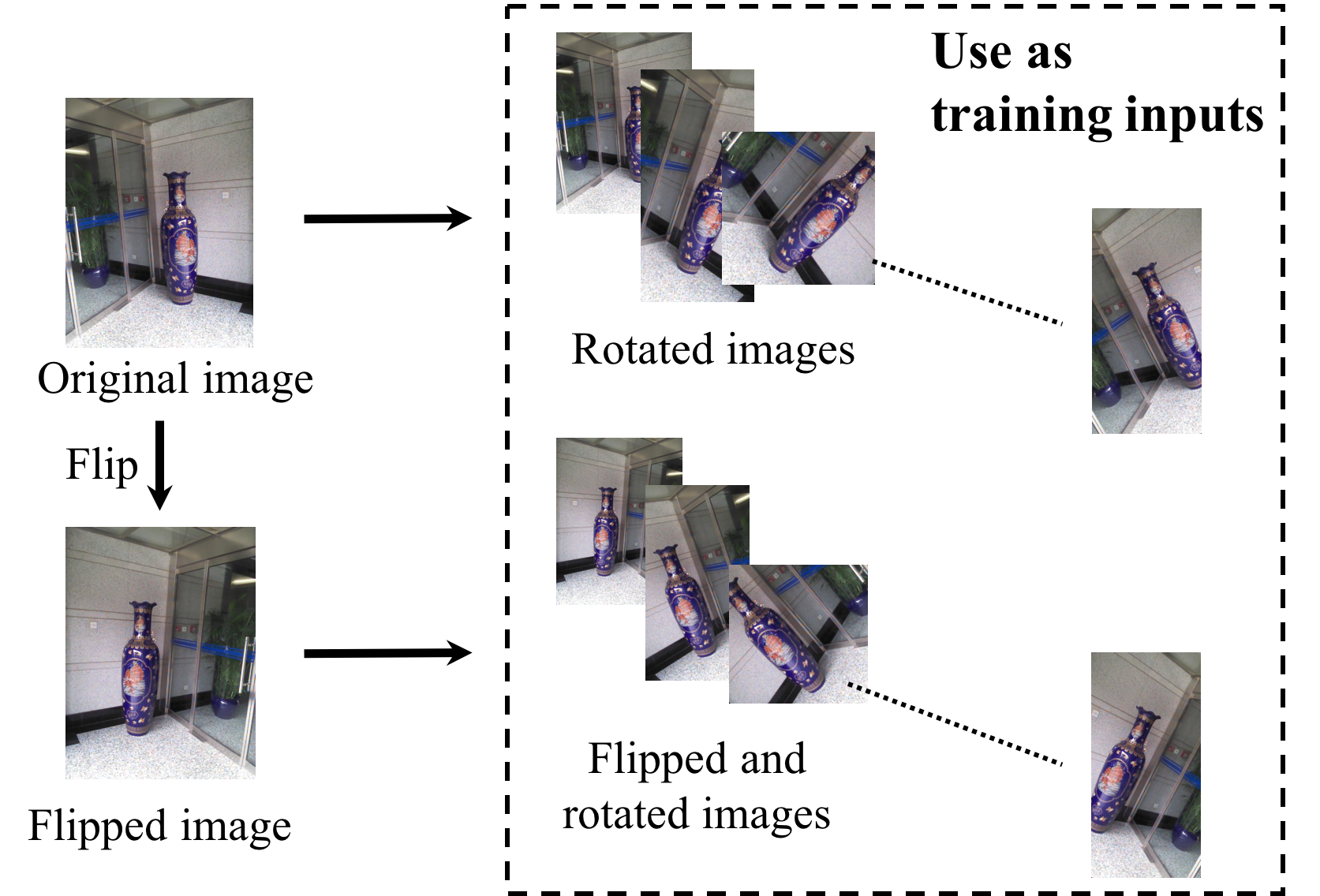}
\end{center}
   \caption{Increasing the number of training datasets by rotating and flipping.}
\label{fig:long}
\label{fig:onecol}
\end{figure}

\begin{figure*}[t]
\begin{center}
\includegraphics[width=0.7\linewidth]{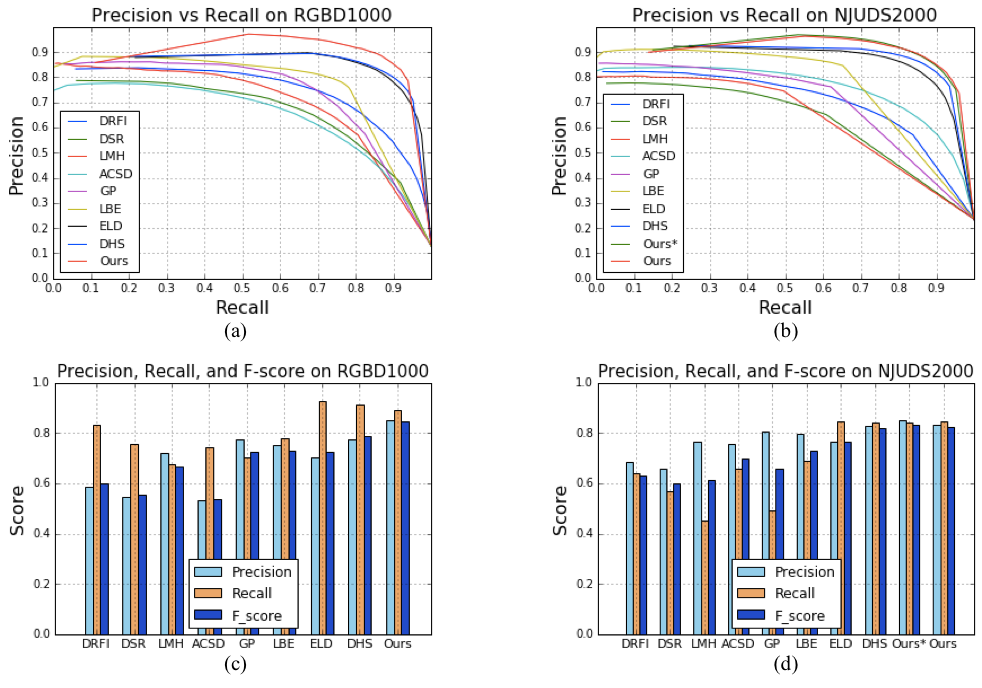}
\end{center}
   \caption{Comparing performance of our methods with other RGB-D saliency methods. The PR curve of our method and the other current RGB-D salient object detection methods on (a) RGBD1000 and (b) NJUDS2000. The F-score of our method and the other current methods on (c) RGBD1000 and (d) NJUDS2000.}
\label{fig:result}
\end{figure*}

\begin{table}
\begin{center}
\begin{tabular}{|c|c|c|}
\hline
 & RGBD1000 & NJUDS2000\\
\hline\hline
DRFI~\cite{Jiang_2013_CVPR} & 0.597689 & 0.631111\\
\hline
DSR~\cite{Li_2013_ICCV} & 0.556269 & 0.599731\\
\hline
LMH~\cite{peng2014rgbd} & 0.667106 & 0.611340\\
\hline
ACSD~\cite{NJUDS} & 0.535048 & 0.696379\\
\hline
GP~\cite{Ren_2015_CVPR_Workshops} & 0.723314 & 0.655915\\
\hline
LBE~\cite{LBE} & 0.727232 & 0.729319\\
\hline
ELD~\cite{Lee_2016_CVPR} & 0.724766 & 0.764561\\
\hline
DHS~\cite{Liu_2016_CVPR} & 0.787486 & 0.817152\\
\hline
{\bf Ours }& {\bf 0.847635} & {\bf 0.821269}\\
\hline
\end{tabular}
\end{center}
\caption{Comparing average F-measure score with other state-of-the-art saliency methods on two datasets.}
\label{table:Fmeasure}
\end{table}

To help address the scarcity of RGB-D salient object datasets, we enhance the training datasets by flipping and rotating images. We made 16 rotated images by rotating the image by 22.5 degree in each step, then each of these is also flipped. As a result, the enhanced training dataset has 32 times as many images as the original. 
For RGBD1000~\cite{peng2014rgbd}, we make 19200 training images from 600 original images. On NJUDS2000~\cite{NJUDS}, we make 38400 training images from the original 1200 images.

The weights for ELD~\cite{Lee_2016_CVPR} can be initialized with a fine-tuned caffemodel. However, this is not suitable for depth, so the weights for depth are initialized randomly. This means the weights for depth need a higher learning rate compared to weights of ELD. We set the learning rate for depth to be 10 times as much as for color. We set our initial base learning rate as 0.05. This means the initial learning rate of depth layers is 0.5. We reduce the learning rate by multiplying by 0.1 in every 10000 learning steps. 1000 superpixels are used for training in every step.

After training with the flipped and rotated datasets, we train our model using only the original images. This is because we assume that the most salient object may change for some images or their saliency maps may become incorrect when the images are flipped or rotated. We use two stage training to mitigate any possible negative effect. In the final original image only training stage, we set the learning rate as 0.01 and do not modify it. We train for 900 steps for RGBD1000~\cite{peng2014rgbd} and 1000 steps for NJUDS2000~\cite{NJUDS}. 1000 superpixels are used for training in every step.

\section{Experimental Evaluation}

We evaluate our architecture's performance on two datasets: RGBD1000~\cite{peng2014rgbd} and NJUDS2000~\cite{NJUDS}. On RGBD1000, we randomly divide the dataset into 600 images for a training set, 200 images for a validation set, and 200 images for test set. On NJUDS2000, we randomly divide the datasets into 1200 images for a training set, 385 images for a validation set, and 400 images for a test set. 

The results are compared against other state-of-the-art RGB-D saliency detection methods: local background enclosure (LBE)~\cite{LBE}; and multi-scale depth-contrast (LMH)~\cite{peng2014rgbd}; and saliency based on region contrast and background, depth, and an orientation prior (GP)~\cite{Ren_2015_CVPR_Workshops}; anisotropic center-surround depth based saliency method (ACSD)~\cite{NJUDS}.
We compare our results also with RGB saliency detection systems: DRFI~\cite{Jiang_2013_CVPR} and DSR~\cite{Li_2013_ICCV} which produce good scores~\cite{2015saliency}.
Finally,
we also add two state-of-the-art CNN-based RGB saliency detection approaches: saliency from low and high level features (ELD)~\cite{Lee_2016_CVPR}; and the Deep hierarchical saliency network (DHS)~\cite{Liu_2016_CVPR}.

\begin{figure*}[tbp]
\begin{center}
\includegraphics[width=0.9\linewidth]{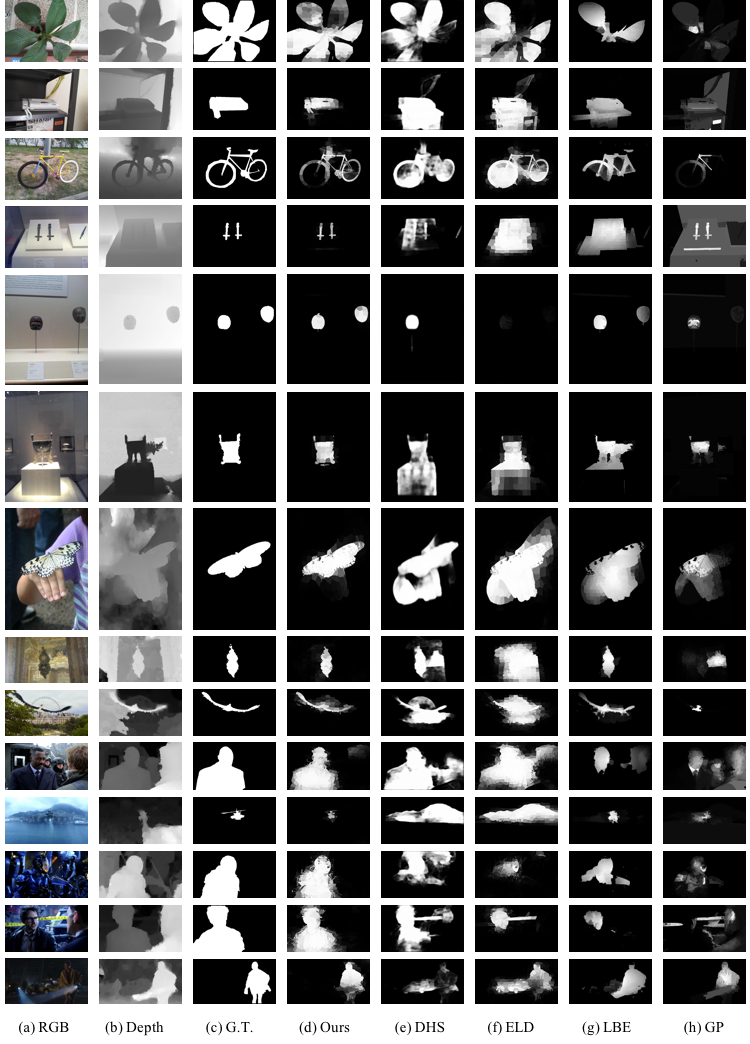}
\end{center}
   \caption{Comparing outputs of our architecture against DHS~\cite{Liu_2016_CVPR}, ELD~\cite{Lee_2016_CVPR}, LBE~\cite{LBE}, GP~\cite{Ren_2015_CVPR_Workshops}. Note that G.T. means Ground Truth.}
\label{fig:final_result}
\end{figure*}

\subsection{Evaluation Criteria}
Like the other state-of-the-art RGB-D salient detection methods~\cite{LBE,peng2014rgbd,Ren_2015_CVPR_Workshops}, we calculate the precision-recall curve and mean F-score for evaluating our results. The F-score is calculated as a following equation:
\begin{equation}
F_\beta =  \frac{(1+\beta^2)Precision \times Recall}{\beta^2 \times Precision + Recall}
\end{equation}
where $\beta=0.3$ in order to put more emphasis on precision than recall~\cite{2009saliency}.

\subsection{Experimental Setup}
As mentioned, we perform training with the datasets augmented with rotated and flipped images, and then train with the original images only. In both cases, we use Adadelta optimizer~\cite{Adadelta} for updating weights. 
For training with the augmented datasets, we set the base learning rate as 0.05, a decay constant $\rho$ as 0.9, and the constant $\epsilon$ as 1e-08. We decrease the base learning in every 10000 iterations by multiplying the base learning rate by 0.1. We perform 50000 training iterations on RGBD1000~\cite{peng2014rgbd} and NJUDS2000~\cite{NJUDS}. 
Then for training with the original images only, we set the base learning rate to 0.01, a decay constant $\rho$ to 0.9, and the constant $\epsilon$ to 1e-08. We perform 900 training iterations on RGBD1000~\cite{peng2014rgbd} and 1000 iterations on NJUDS2000~\cite{NJUDS}.
These parameter values were determined by performance on validation datasets. 

\subsection{Results}

Our learning architecture outperforms the other RGB-D salient object detection methods (Figure 8a and 8b, Table \ref{table:Fmeasure}). Our method is particularly effective for high recall rates with respect to other methods.
Our approach outperforms the results of bottom-up approaches such as LBE~\cite{LBE} and LMH~\cite{peng2014rgbd} (Figure 8a and 8b). In addition, compared to other top-down RGB salient object detection systems such as ELD-Net~\cite{Lee_2016_CVPR} and DHSNet~\cite{Liu_2016_CVPR}, our approach performs better on the P-R curve and F-score. 

\begin{table}
\begin{center}
\begin{tabular}{|c|c|c|c|}
\hline
 & Precision & Recall & F-measure\\
\hline\hline
Ours & 0.834091 & 0.843668 & 0.821269\\
\hline
with mean depth & 0.850724 & 0.840648 & 0.833303\\
\hline
\end{tabular}
\end{center}
\caption{Replacing the superpixel histogram with mean depth improves results for NJUDS2000~\cite{NJUDS} where depth data is noisy.}
\label{NJUDShisto}
\end{table}

On the NJUDS2000~\cite{NJUDS}, we perform training without using $\chi^2$ distance of histogram difference of the depth of the superpixel and grid cells, and using the average depth of the superpixel instead. This is because the quality of the depth images is not as good on NJUDS2000 datasets, as the depth images are captured by stereo camera.  This change leads to an improvement in performance. (Figure 8b and 8d, Table \ref{NJUDShisto}) We name this method as Ours* in Figure \ref{fig:result}. In general, this may be an effective approach if training data has noisy depth.

\begin{table}
\begin{center}
\begin{tabular}{|c|c|c|c|}
\hline
 & Precision & Recall & F-measure\\
\hline\hline
RGB only (ELD) & 0.700276 & 0.927371 & 0.724766\\
\hline
Without BED & 0.840953 & 0.891432 & 0.840704\\
\hline
Ours & 0.84833 & 0.890801 & 0.847635\\
\hline
\end{tabular}
\end{center}
\caption{Comparing scores with different input features on RGBD1000~\cite{peng2014rgbd}.}
\label{RGBD_Discussion}
\end{table}

\begin{table}
\begin{center}
\begin{tabular}{|c|c|c|c|}
\hline
 & Precision & Recall & F-measure\\
\hline\hline
RGB only (ELD) & 0.766507 & 0.844895 & 0.764561\\
\hline
Without BED & 0.830839 & 0.841835 & 0.816602\\
\hline
Ours & 0.834091 & 0.843668 & 0.821269\\
\hline
\end{tabular}
\end{center}
\caption{Comparing scores with different input features on NJUDS2000~\cite{NJUDS}.}
\label{NJUDS_Discussion}
\end{table}

In order to evaluate the effect of the BED features, we measure the performance of our methods without using BED features. We perform training in the same architecture other than BED features, perform the same training, and use the same measures of performance. Tables \ref{RGBD_Discussion} and \ref{NJUDS_Discussion} shows the results. The tables contain average precision, recall, and F-measure of three methods, the result of ELD-Net~\cite{Lee_2016_CVPR}, our network without using six BED features, and our architecture. As can be seen BED contributes to an increase in the results. On the RGBD1000 dataset, precision increases well while holding the same recall. On NJUDS2000 datasets, precision increases and recall rate also increases slightly. 

Figure \ref{fig:final_result} shows the output of our architecture with the other state-of-the-art methods. 

\section{Conclusion}
In this paper, we proposed a novel architecture that provides the first complete RGB-D salient object detection systems using a deep CNN. We incorporate a novel mid-level feature, BED, to capture background enclosure, as well as low level depth cues that incorporate depth contrast, and high level features.
Our results demonstrate that our novel architecture outperforms other RGB-D salient object detection methods. Further, we show that adding low-level depth and BED each yield an improvement to the detection results.

{\small
\bibliographystyle{ieee}
\bibliography{egpaper_final}
}

\section{Supplementary material}

\subsection{Introduction}
The purpose of this supplementary material is to analyze the results of our method more closely. First, we examine cases where our method succeeds in reducing false positives compared to: the other state-of-the-art RGB learning based saliency methods, DHSNet~\cite{Liu_2016_CVPR} and ELDNet~\cite{Lee_2016_CVPR}; and RGB-D bottom-up saliency methods, LBE~\cite{LBE} and GP~\cite{Ren_2015_CVPR_Workshops}. An increase of the precision rate is shown on the Figure \ref{fig:result}c and \ref{fig:result}d. This means our architecture is better able to reduce false positives compared to previous state-of-the-art methods. We also show cases where our method succeeds in reducing false negatives compared to DHS~\cite{Liu_2016_CVPR}, ELD~\cite{Lee_2016_CVPR}, LBE~\cite{LBE}, and GP~\cite{Ren_2015_CVPR_Workshops}. Then, in order to analyze the performance of the BED features we compare the output with that when BED features are not incorporated. Finally, we illustrate our failure cases and analyze the reasons. 

Please find the attached all test outputs of our method, LBE~\cite{LBE}, DHS~\cite{Liu_2016_CVPR}, and GP~\cite{Ren_2015_CVPR_Workshops}. The outputs of LBE~\cite{LBE}, DHS~\cite{Liu_2016_CVPR}, and GP~\cite{Ren_2015_CVPR_Workshops} are obtained by running code from the authors' websites.

\subsection{Comparing our results with other methods.}

\begin{figure*}[t]
\begin{center}
\includegraphics[width=0.9\linewidth]{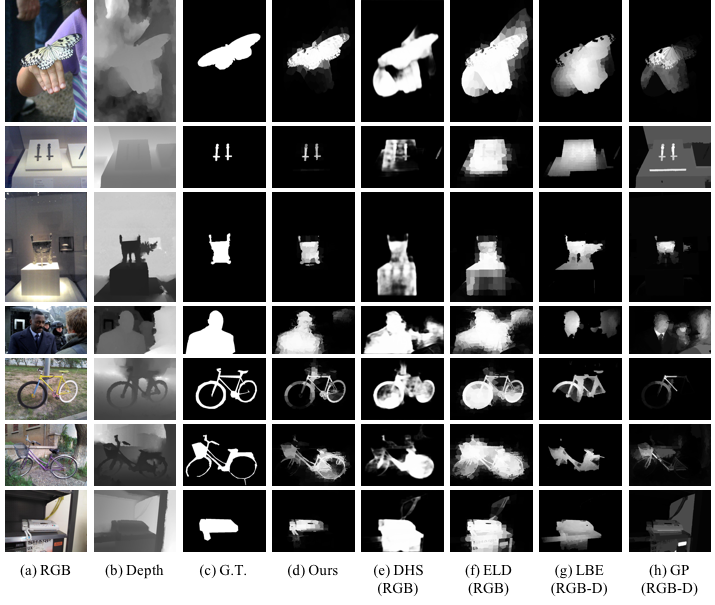}
\end{center}
   \caption{Cases where our method succeeds in reducing false positives compared to state-of-the-art methods, DHS~\cite{Liu_2016_CVPR}, ELD~\cite{Lee_2016_CVPR}, LBE~\cite{LBE}, GP~\cite{Ren_2015_CVPR_Workshops}. Note that RGB under the name represents RGB learning based methods, and RGB-D represents RGB-D bottom-up approach methods. G.T. means Ground Truth.}
\label{fig:FalsePositives}
\end{figure*}

\begin{figure*}[t]
\begin{center}
\includegraphics[width=0.9\linewidth]{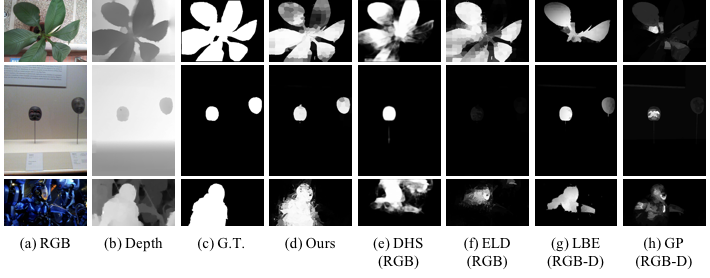}
\end{center}
   \caption{Cases where our method succeeds in reducing false negatives compared to other state-of-the-art methods, DHS~\cite{Liu_2016_CVPR}, ELD~\cite{Lee_2016_CVPR}, LBE~\cite{LBE}, GP~\cite{Ren_2015_CVPR_Workshops}. G.T. means Ground Truth.}
\label{fig:FalseNegatives}
\end{figure*}

As can be seen in Figure \ref{fig:result}c and \ref{fig:result}d, a strong point of our method is that it increases precision while maintaining a high recall rate compared to other state-of-the-art RGB salient object detection methods~\cite{Lee_2016_CVPR,Liu_2016_CVPR}, and has a higher precision and recall rate than other RGB-D state-of-the-art salient object detection methods~\cite{LBE,Ren_2015_CVPR_Workshops}. This means our method is better able to exclude false positives. We provide such examples in Figure \ref{fig:FalsePositives}.  In these examples, RGB top-down methods are able to detect the correct regions, however, these methods also detect many false positives. By combining color and depth data, our method solves this problem. In Figure \ref{fig:FalsePositives} row 1, for example, the salient regions are difficult to detect without knowledge about the butterfly. Thus, the other state-of-the-art RGB and RGB-D methods fail to detect the region, while our method succeeds in detecting the insect utilizing both the color and depth images. A similar output is also shown in Figure \ref{fig:FalsePositives} row 2. Though state-of-the-art RGB salient object detection methods fail and the salient regions are difficult to detect from the depth map, our method succeeds in detecting the salient regions combining color and depth features. In Figure \ref{fig:FalsePositives} row 3-7, our method succeeds in detecting the salient objects. Though the other RGB top-down methods succeed in detecting the salient regions, they also find false positives. On these examples, other RGB-D bottom-up methods detect many false positives and negatives. 

As mentioned above, our method is high capable of excluding false positives, but our method also helps to reduce false negatives. We illustrate such examples in Figure \ref{fig:FalseNegatives}. For all examples, other state-of-the-art RGB and RGB-D methods fail to detect the entire salient regions. For example, in Figure \ref{fig:FalseNegatives} row 1, the other methods fail to detect all leaves, while our method succeeds by utilizing color and depth features. 

\subsection{Comparing our outputs including and not including BED features.}

\begin{figure*}[t]
\begin{center}
\includegraphics[width=0.9\linewidth]{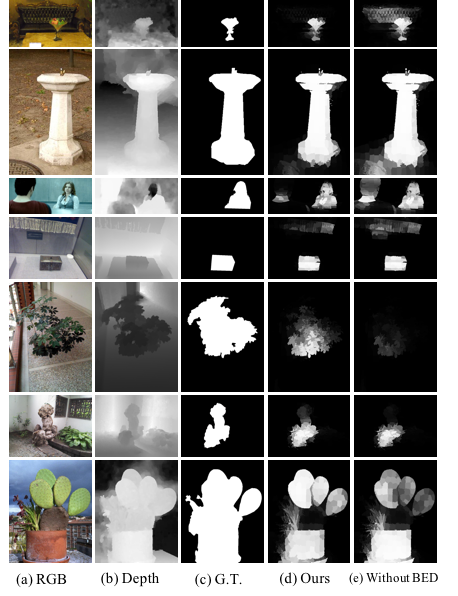}
\end{center}
   \caption{Comparing the results with BED features and without BED features. BED features help to reduce false positives and false negatives in some cases. G.T. means Ground Truth.}
\label{fig:BED_Examples}
\end{figure*}

For further analysis of the effectiveness of the BED features, we compare the outputs using our full system versus our system with BED features excluded. 
Figure \ref{fig:BED_Examples} shows examples where the BED features lead to improved performance.

BED features help to reduce false positives in some cases, for example, see Figure \ref{fig:BED_Examples} row 1-4. Figure \ref{fig:BED_Examples} row 1 shows this effect clearly. By making use of the BED features, our method succeeds in excluding the chair at the back and reduces false positives. BED features also help to reduce false negatives in some images, for example, see Figure \ref{fig:BED_Examples} row 5-7. In Figure \ref{fig:BED_Examples} row 5, our method is more sucessful at detecting the salient objects thanks to the BED features. Similar effects can seen in Figure \ref{fig:BED_Examples} row 6 and 7.

\subsection{Failure cases.}

\begin{figure*}[t]
\begin{center}
\includegraphics[width=0.9\linewidth]{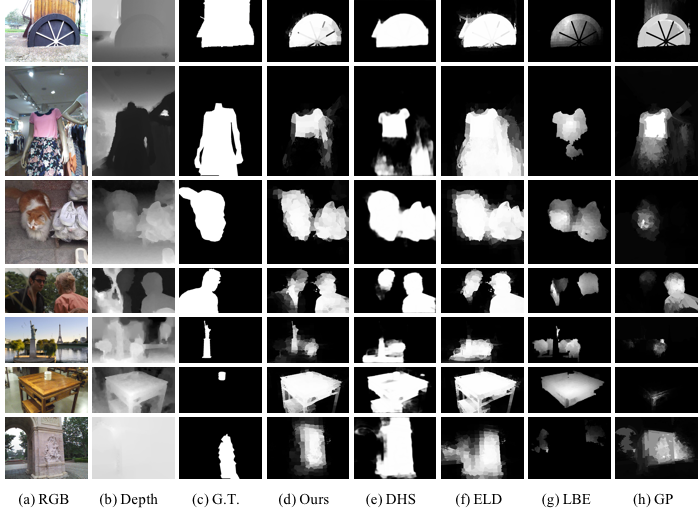}
\end{center}
   \caption{Examples of failure cases of our architecture. We also show the outputs of other state-of-the-art methods, DHS~\cite{Liu_2016_CVPR}, ELD~\cite{Lee_2016_CVPR}, LBE~\cite{LBE}, GP~\cite{Ren_2015_CVPR_Workshops}. Our method fails to detect salient regions when particular knowledge is needed, such as what is wheel or role of face direction in attention. G.T. means Ground Truth.}
\label{fig:failure_cases}
\end{figure*}

Finally, we include examples of failure cases of our method in Figure \ref{fig:failure_cases}. At first sight of the image in Figure \ref{fig:failure_cases} row 1, we may consider that only the wheel is a salient object. However,  actually the wheel is connected with the square pole in the back, which is also labeled as salient in this dataset. In such a case, it is difficult to detect the whole salient object without knowledge of the object. In Figure \ref{fig:failure_cases} row 2, we can see a similar situation. At first sight, we may think only the pink T-shirt is a salient region, but actually the skirt is also labeled as  salient. Humans can detect both of them as salient objects because we know both T-shirts and skirts are clothes, but our algorithm does not. In Figure \ref{fig:failure_cases} row 3, humans label only the cat as a salient object because it is an animal and we usually pay more attention to animals than non-living objects such as shoes. However, for an algorithm it is difficult to detect only the cat without such knowledge. In Figure \ref{fig:failure_cases} row 4, saliency is strongly related with face direction. The man on the left attracts more attention because is he is looking more towards the camera. Our algorithm seems not to consider this fact. Saliency in Figure \ref{fig:failure_cases} row 5 is related to the face direction of the statue and the fact that humans usually pay more attention to statues than trees. The center tree is also detected as a salient object by our algorithm, we believe, because it seems not to have such knowledge. Figure \ref{fig:failure_cases} row 6 is a difficult case because the salient object is very small and prerequisite knowledge that an object on the desk is more important than the desk is needed.  Figure \ref{fig:failure_cases} row 7 is a case where both color and depth features do not seem to be sufficient detecting the salient regions. Saliency here is quite subtle. Our method can detect similar regions in this case but not clearly.

\end{document}